# Pectoral Muscles Suppression in Digital Mammograms using Hybridization of Soft Computing Methods

I. Laurence Aroquiaraj and K. Thangavel

*Abstract*— Breast region segmentation is an essential prerequisite in computerized analysis of mammograms. It aims at separating the breast tissue from the background of the mammogram and it includes two independent segmentations. The first segments the background region which usually contains annotations, labels and frames from the whole breast region, while the second removes the pectoral muscle portion (present in Medio Lateral Oblique (MLO) views) from the rest of the breast tissue. In this paper we propose hybridization of Connected Component Labeling (CCL), Fuzzy, and Straight line methods. Our proposed methods worked good for separating pectoral region. After removal pectoral muscle from the mammogram, further processing is confined to the breast region alone. To demonstrate the validity of our segmentation algorithm, it is extensively tested using over 322 mammographic images from the Mammographic Image Analysis Society (MIAS) database. The segmentation results were evaluated using a Mean Absolute Error (MAE), Hausdroff Distance (HD), Probabilistic Rand Index (PRI), Local Consistency Error (LCE) and Tanimoto Coefficient (TC). The hybridization of fuzzy with straight line method is given more than 96% of the curve segmentations to be adequate or better. In addition a comparison with similar approaches from the state of the art has been given, obtaining slightly improved results. Experimental results demonstrate the effectiveness of the proposed approach.

## I. INTRODUCTION

Breast cancer is the most common of all cancers and is the leading cause of cancer deaths in women worldwide, accounting for more than 1.6% of deaths and case fatality rates are highest in low-resource countries. A recent study of breast cancer risk in India revealed that 1 in 28 women develop breast cancer during her lifetime. This is higher in urban areas being 1 in 22 in a lifetime compared to rural areas where this risk is relatively much lower being 1 in 60 women developing breast cancer in their lifetime. In India the average age of the high risk group in India is 43-46 years unlike in the west where women aged 53-57 years are more prone to breast cancer. As there is no effective method for its prevention, the diagnosis of breast cancer in its early stage of development has become very crucial for the prevention of cancer. Computer-aided diagnosis (CAD) systems play an important role in earlier diagnosis of breast cancer [1].

Every effort directed to improve early detection is needed. Most of them require an initial processing step that splits the image into interesting areas, such as the breast region, background and patient markings (artifacts).

Mammograms capture the low energy X-rays which passes through a compressed breast. It is used to detect tumors and cysts in an advanced stage of cancer and to help distinguish benign (noncancerous) and malignant (cancerous) cases.

Pectoral muscles are the regions in mammograms that contain brightest pixels. These regions must be removed before detecting the tumor cells so that mass detection can be done efficiently. Pectoral muscles lie on the left or right top corner depending on the view of the image. We must detect the position of the pectoral muscles (left top corner or right top corner), before removing it. Remove the pectoral muscle from the mammogram images, was a more complicated task because of different dimensions, shapes and pixel intensities from the muscles in the images. The Hough transform was used, a commonly used technique to detect geometrical shapes in images. In our case, we wanted to detect the hybridization of CCL, Fuzzy and Straight Line methods that separates the muscle from the breast and apply a mask to keep the image without it. It was also necessary to apply some pre-processing procedures to the images like smoothing, finding the orientation, extract a region of interest that contains the pectoral muscle.

The rest of the paper is organized as follows: Section 2 presents the research back ground and ccl, straight line and fuzzy methods. Section 3 presents the mammogram image data acquisition. Section 4 describes the proposed hybridization of the three methods. The experimental results are discussed in section 5 and conclusion is presented in section 6.

## II. RESEARCH BACKGROUND

There have been various approaches to the task of segmenting the pectoral muscle. A histogram-based thresholding technique is used by K.Thangavel et. al. [2] to separate the pectoral muscle region. For selecting the threshold value the global optimum is considered. The

I. Laurence Aroquiaraj was IEEE Member, Assistant Professor, Department of Computer Science, Periyar University, Salem – 636 011, Tamil Nadu, India (corresponding author to provide phone: +918807058505; e-mail: laurence.raj@gmail.com).
Prof. Dr. K. Thangavel, was IEEE Member, Professor and Head, Department of Computer Science, Periyar University, Salem – 636 011, Tamil Nadu, India (e-mail: drktvelu@yahoo.com).

intensity values smaller than global optimum threshold are changed to zero, and the gray values greater than the threshold are changed to one. To better preserve the pectoral muscle region erosion and dilation operations are applied. To segment the pectoral muscle region the gray level mammogram image is converted to binary image. The white pixels in the lower left corner of the mammogram image indicate the pectoral muscle region. The segmentation outputs of these methods were very efficient and excellent.

K.Thangavel et. al. proposed in [2] applies the meta-heuristic methods such as Ant Colony Optimization (ACO) and Genetic Algorithm (GA) for identification of suspicious region in mammograms.

M. Wirth et al. developed an algorithm [3] that uses morphological preprocessing and fuzzy rule-based algorithm for breast region extraction.

Kostas Marias et al. [4] used the boundary extraction technique based on a combination of the Hough transform followed by image gradient operators and morphology in order to make coherent the breast region part of the image.

Gradient based method is proposed by Méndez et al. [5] to find the breast contour. They used a two level thresholding technique to isolate the breast region of the mammogram. The smoothed mammogram is divided into three regions and then a tracking algorithm is applied to the mammogram to detect the border.

T.S.Subashini et al. [6] proposed a technique for pectoral muscle removal and detecting masses in mammograms using connected component labeling (CCL). From the detected corner pixel the intensity discontinuity is detected on each and every column of the same row. Coordinates of the pixel in which the intensity change is encountered is considered as width of the pectoral region.

Arun Kumar et al. [7] proposed wavelet decomposition with edge detection using canny filter.They used inside the pectoral muscle region which removed by approximating muscle boundary with a straight-line that connects upper right corner and lower left corner of muscle region in the case of the right breast image.

Lou et al. [8] proposed a method based on the assumption that the trace of intensity values from the breast region to the air-background is a monotonic decreasing function. One of the inherent limitations of these methods is the fact that very few of them preserve the skin or nipple.

Zhili Chen et al. [9] proposed a fully automated breast region segmentation method based on histogram thresholding, edge detection in scale space, contour growing and polynomial fitting. Subsequently, pectoral muscle removal using region growing is presented.

Roshan Dharshana Yapa et al. presented a new algorithm [10] for estimating skin-line and breast segmentation using fast marching algorithm. They introduced some modifications to the traditional fast marching method, specifically to improve the accuracy of skin-line estimation and breast tissue segmentation.

The most promising method of extracting the breast contour focuses on modeling the non-breast region of a mammogram using a polynomial method, as described by Chandrasekhar and Attikiouzel [11, 12].

Kwork et al. [13] developed a method for automatic pectoral muscle segmentation on mammograms by straightline estimation and cliff detection. A straight line estimates the muscle edge and cliff detection refines the detected edge by surface smoothing and edge detection in a restricted neighborhood.

H. Mirzaalian et al. developed [14] a new method for the identification of the pectoral muscle in MLO mammograms. The developed method is based on nonlinear diffusion algorithm. They compared their results by those recognized by two expert radiologists. To evaluate the accuracy of proposed method, HDM (Hausdorff Distance Measure) and MAEDM (Mean of Absolute Error Distance Measure) were used.

R.J. Ferrari proposed [15] a new method for the identification of the pectoral muscle in MLO mammograms based upon a multi resolution technique using Gabor wavelets. This new method overcomes the limitation of the straight-line representation considered in their initial investigation. The results of the Gabor-filter-based method indicated low Hausdorff distances with respect to the hand-drawn pectoral muscle edges.

Mario Mustra et al. [16] uses wavelet decomposition, image blurring and edge detection using the Sobel filter for breast border detection and pectoral muscle segmentation.

N.Nicolau et al. [17] proposed the use of Independent Component Analysis (ICA) for identification and sub sequent removal of the pectoral muscle.

### III. PROPOSED WORK

#### A. Connected Components Labeling (CCL)

Connected-component labeling (alternatively connected-component analysis, blob extraction, region labeling, blob discovery, or region extraction) is an algorithmic application of graph theory, where subsets of *connected components* are uniquely *labeled* based on a given heuristic. Connected-component labeling is not to be confused with segmentation.

Connected-component labeling is used in computer vision to detect connected regions in binary digital images, although color images and data with higher- dimensionality can also be processed. When integrated into an image recognition system or human-computer interaction interface, connected component labeling can operate on a variety of information. Blob extraction is generally performed on the resulting binary image from a thresholding step. Blobs may be counted, filtered, and tracked. [6]

> Algorithm: **Connected Components Labeling**
> Step 1: Iterate through each element of the data by column, then by row (Raster Scanning)
> Step 2: If the element is not the background
>     Step 2.1: Get the neighboring elements of the current element
>     Step 2.2: If there are no neighbors, uniquely label the current element and continue
>     Step 2.3: Otherwise, find the neighbor with the smallest label and assign it to the current element
>     Step 2.4: Store the equivalence between neighboring labels
> Step 3: Iterate through each element of the data by column, then by row
> Step 4: If the element is not the background
> Step 5: Relabel the element with the lowest equivalent label

*B. Fuzzy Logic Method*

**Fuzzy logic** is a form of many-valued logic; it deals with reasoning that is fixed or approximate rather than fixed and exact. In contrast with "crisp logic", where binary sets have two-valued logic: true or false, fuzzy logic variables may have a truth value that ranges in degree between 0 and 1. Fuzzy logic has been extended to handle the concept of partial truth, where the truth value may range between completely true and completely false. Furthermore, when linguistic variables are used, these degrees may be managed by specific functions [18].

Fuzzy set theory defines fuzzy operators on fuzzy sets. The problem in applying this is that the appropriate fuzzy operator may not be known. For this reason, fuzzy logic usually uses IF-THEN rules, or constructs that are equivalent, such as fuzzy associative matrices.

Rules are usually expressed in the form: IF *variable* IS *property* THEN *action*

> **Algorithm: Fuzzy Logic Method**
> Step 1: Iterate through each element of the data by column, then by row (Raster Scanning)
> Step 2: Fuzzifying the given image based on expert knowledge using fuzzy logic
> Step 3: Modifying the membership values of fuzzified image by using fuzzy techniques such as fuzzy clustering, fuzzy rule based approach etc.
> Step 4: Defuzzifying the modified membership values of the image
> Step 5: Getting the resultant image

*C. Straight line method*

In Straight line method, the algorithm detects the straight-line that represents the boundary of pectoral muscle in mammogram images. After the region is selected using the histogram based method, the x- axis and y- axis values will be found for the pectoral region. Then depending up on the (x, y) and (x', y') the straight line will be drawn in the index value of the region. The straight line will be drawn using around half as Cuba's coding process [13].

> **Algorithm: Straight Line Method**
> Step 1: Getting the 3 * 3 (Raster Scanning)
> Step 2: Each center of that three values will be taken as 1's
> Step 3: Next diagonal for the first two will be taken for 3*3 Matrices
> Step 4: The process will be continuing depending upon the (x, y) and (x', y') the straight line will be drawn in the index value of the region.
> Step 5: Getting the resultant image

*D. Statistical Measurement*

*1. Probabilistic Rand Index*

The section introduces a measure that combines the desirable statistical properties of the Rand index with the ability to accommodate refinements appropriately. Since the latter property is relevant primarily when quantifying consistency of image segmentation results, we will focus on that application while describing the measure.

Consider a set of manually segmented (ground truth) images $\{S_1, S_2, \dots, S_k\}$ corresponding to an image X = $\{x_1, x_2, \dots, x_i, \dots, x_N\}$ where a subscript indexes one of N pixels. Let S be the segmentation that is to be compared with the manually labeled set. Our goal is to compare a candidate segmentation S to this set and obtain a suitable measure of consistency $d(S, S_{1 \dots K})$. Given the manually labeled images, we can compute the empirical probability of the label relationship of a pixel pair $x_i$ and $x_j$ simply as:

$$\hat{P}(\hat{l}_i = \hat{l}_j) = \frac{1}{K} \sum_{k=1}^{K} I(l_i^{(k)} = l_j^{(k)})$$

$$\hat{P}(\hat{l}_i \neq \hat{l}_j) = \frac{1}{K} \sum_{k=1}^{K} I(l_i^{(k)} \neq l_j^{(k)})$$

$$= 1 - \hat{P}(\hat{l}_i = \hat{l}_j)$$

Consider the probabilistic Rand (PR) index:

$$PR(S, S_{\{1,\dots,K\}}) = \frac{1}{\binom{N}{2}} \sum_{i \neq j} [\mathbf{I}(l_i = l_j) \hat{P}(\hat{l}_i = \hat{l}_j) + \mathbf{I}(l_i \neq l_j) \hat{P}(\hat{l}_i \neq \hat{l}_j)] \quad (1)$$

This measure takes values in [0, 1] – 0 when S and $\{S_1, S_2 \ldots S_K\}$ have no similarities (i.e. when S consists of a single cluster and each segmentation in $\{S_1, S_2, \ldots, S_K\}$ consists only of clusters containing single points, or vice versa) to 1 when all segmentations are identical.

## 2. Local Consistency Error

Let $S_1$, $S_2$ are two segmentations, $R_{1,i}$ is the set to a region of pixels corresponding in the $S_1$ segmentation and containing the pixel I, |R| is the set cardinality and | is the set difference. A refinement tolerant measure error was defined at each pixel i:

$$\varepsilon_i(S_1, S_2) = \frac{|R_{1,i} \setminus R_{2,i}|}{|R_{1,i}|} \quad (2)$$

This non-symmetric local error measure encodes a measure of refinement in one direction only. Local Consistency Error (LCE) allows refinement in both directions.

$$LCE(S_1, S_2) = \frac{1}{n} \sum_i \min\{\varepsilon_i(S_1, S_2), \varepsilon_i(S_2, S_1)\} \quad (3)$$

## 3. Tanimoto Coefficient

The use of Tanimoto coefficient has become popular as a coefficient of similarity in images.

The Tanimoto Coefficient is an extended Jaccard Coefficient. The Jaccard similarity coefficient is a static used for comparing the similarity and diversity of sample sets. The Jaccard coefficient is defined as the size of the intersection divided by the size of the union of the sample sets:

$$J(A,B) = \frac{|A \cap B|}{|A \cup B|} \quad (4)$$

The Jaccard distance, which measures dissimilarity between sample sets, is obtained by dividing the difference of the sizes of the union abd the intersection of two sets by the size of the union, or simpler, by subtracting the Jaccard coefficient from 1 as done in the following equation.

$$J\delta(A,B) = 1 - J(A,B) = \frac{|A \cup B| - |A \cap B|}{|A \cup B|} \quad (5)$$

For two objects, A and B, each with n binary attributes, the Jaccard coefficient is a useful measure of the overlap that A and B share with their attributes. Each attribute of A and B can either be 0 or1. The total number of each combination of attributes for both A and B are specified as follows:

The Jaccard similarity coefficient J is given by the following equation:

$$J = \frac{M11}{M01 + M10 + M11}$$

The Jaccard distance, J', is given by the following equation:

$$J' = \frac{M01 + M10}{M01 + M10 + M11}$$

Where,

**M00** – Represents the total number of attributes where both A and B have value of 0.

**M01** – Represents the total number of attributes where the attribute of A is 0 and the attribute of B is1

**M10** - Represents the total number of attributes where the attribute of A is 0 and the attribute of B is1

**M11** - Represents the total number of attributes where both A and B have value of 1.

**M00+M01+M10+M11** – Each attribute has fall into these four categories.

Cosine similarity is a measure of similarity between two vectors of n dimensions by finding the angle between them, often used to compare documents in text mining. Given two vectors of attributes, A and B, the cosine similarity, $\theta$, is represented using a dot product and magnitude as in the following equation:

$$\theta = arc(\cos)\frac{A,B}{\|A\|\|B\|}$$

Since the angle, $\theta$, is in the range of $[0, \pi]$, the resulting similarity will yield the value of $\pi$ meaning exactly the opposite, $\pi/2$ meaning independent, 0 meaning exactly the same and in-between values indicating intermediate similarities or dissimilarities.

This cosine similarity metric may be extended such that it yields the Jaccard coefficient in the case of binary attributes and can be represented as the Tanimoto coefficient represented in the below equation.

$$T(A,B) = \frac{A,B}{\|A\|^2 + \|B\|^2 - A,B} \quad (6)$$

In some case, each attribute is binary such that each bit represents the absence of presence of a characteristic, thus, it is better to determine the similarity via the overlap, or intersection, of the sets.

Simply put, the Tanimoto Coefficient uses the ratio of the intersecting set to the union set as the measure of similarity. Represented as a mathematical equation:

$$T_a = \frac{N_c}{N_a + N_b - N_c} \quad (7)$$

In this equation, N represents the number of attributes in each object (a, b). C in this case is the intersection set.

## 4. Mean absolute error

In statistics, the mean absolute error (MAE) is a quantity used to measure how close forecasts or predictions are to the eventual outcomes. The mean absolute error is given by

$$MAE = \frac{1}{n}\sum_{i=1}^{n}|f_i - y_i| = \frac{1}{n}\sum_{i=1}^{n}|e_i| \quad (8)$$

As the name suggests, the mean absolute error is an average of the absolute errors $e_i = f_i - y_i$ where $f_i$ is the prediction and $y_i$ the true value. Note that alternative formulations may include relative frequencies as weight factors.

The mean absolute error is a common measure of forecast error in time series analysis, where the terms "mean absolute deviation" is sometimes used in confusion with the more standard definition of mean absolute deviation. The same confusion exists more generally.

*5. Hausdorff distance measures*

Hausdorff distance, measures how far two subsets of a metric space are from each other. It turns the set of non-empty compact subsets of a metric space into a metric space in its own right. Informally, two sets are close in the Hausdorff distance if every point of either set is close to some point of the other set. The Hausdorff distance is the longest distance you can be forced to travel by an adversary who chooses a point in one of the two sets, from where you then must travel to the other set. In other words, it is the farthest point of a set that you can be to the closest point of a different set.

Let *X* and *Y* be two non-empty subsets of a metric space $(M, d)$. We define their Hausdorff distance $d_H(X, Y)$ by

$$d_H(X,Y) = \max\left\{\sup_{x \in X} \inf_{y \in Y} d(x,y), \sup_{y \in Y} \inf_{x \in X} d(x,y)\right\} \quad (9)$$

Where *sup* represents the supremum and *inf* the infimum.

Equivalently $d_H(X,Y) = \inf\{\varepsilon > 0; X \subseteq Y_\varepsilon \text{ and } Y \subseteq X_\varepsilon\}$

Where $X_\varepsilon = \bigcup_{x \in X}\{z \in M; d(z,x) \leq \varepsilon\}$

That is, the set of all points within ε of the set *X* (sometimes called a generalized ball of radius ε around *X*).

We can create a metric by defining the Hausdorff distance to be:

$$d_H(X,Y) = \max\{d(X,Y), D(Y,X)\} \quad (10)$$

IV. RESULTS AND DISCUSSIONS

Obtaining real mammogram images (322 images) for carrying out research is highly difficult due to privacy issues, legal issues and technical hurdles. Hence the Mammography Image Analysis Society (**MIAS**) database (ftp://peipa.essex.ac.uk) is used in this paper to study the efficiency of the proposed pectoral muscle removal image segmentation and evaluated using mammography images. The pectoral muscle region removal from the mammogram is had by our three proposed method. Six methods were Connected Component Labeling (CCL), Fuzzy, Straight line, CCL with Fuzzy, CCL with Straight line, and Straight line with Fuzzy. These six methods were compared with each other by error measure, index measure and distance measure.

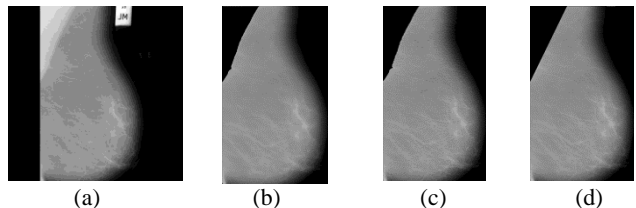

(a) (b) (c) (d)

Figure 1 (a) Pectoral muscle identification results for MIAS image mdb006
(b) CCL with Fuzzy
(c) CCL with Straight Line
(d) Fuzzy with Straight Line

Figure 1 shows the breast region of left and right mammograms after the removal of pectoral muscle region using three different methods like ccl with fuzzy, ccl with straight line and fuzzy with straight line.

It is observed from the Table 1 that proposed methods produces minimized probabilistic rand index values. Figure 2 shows a comparison of all six methods average values of ten mammogram images obtained by using the proposed methods

TABLE I
PROBABILISTIC RAND INDEX (PRI) VALUES FOR CCL, FUZZY, STRAIGHT LINE AND PROPOSED METHOD

| MIAS Images | CCL | FUZZY | St. Line |
|---|---|---|---|
| Mdb006 | 0.72017 | 0.719915 | 0.718702 |
| Mdb015 | 0.597211 | 0.5977 | 0.596224 |
| Mdb020 | 0.730559 | 0.730677 | 0.729346 |
| Mdb040 | 0.4382 | 0.439102 | 0.419749 |
| Mdb042 | 0.56572 | 0.566003 | 0.56086 |
| Mdb050 | 0.757102 | 0.757341 | 0.755594 |
| Mdb070 | 0.68304 | 0.683109 | 0.680662 |
| Mdb077 | 0.690741 | 0.690677 | 0.690618 |
| Mdb099 | 0.549994 | 0.550095 | 0.548686 |
| Mdb100 | 0.583248 | 0.583436 | 0.578468 |

| MIAS Images | CCL with FUZZY | CCL with St. Line | FUZZY with St. Line |
|---|---|---|---|
| Mdb006 | 0.719722 | 0.720041 | 0.71951 |
| Mdb015 | 0.563394 | 0.563353 | 0.563298 |
| Mdb020 | 0.664296 | 0.665386 | 0.665227 |
| Mdb040 | 0.44328 | 0.443519 | 0.435582 |
| Mdb042 | 0.547105 | 0.547259 | 0.545755 |
| Mdb050 | 0.638295 | 0.638418 | 0.638398 |
| Mdb070 | 0.610763 | 0.61171 | 0.611678 |
| Mdb077 | 0.689599 | 0.689809 | 0.689286 |
| Mdb099 | 0.57717 | 0.576323 | 0.593367 |
| Mdb100 | 0.593833 | 0.59333 | 0.593367 |

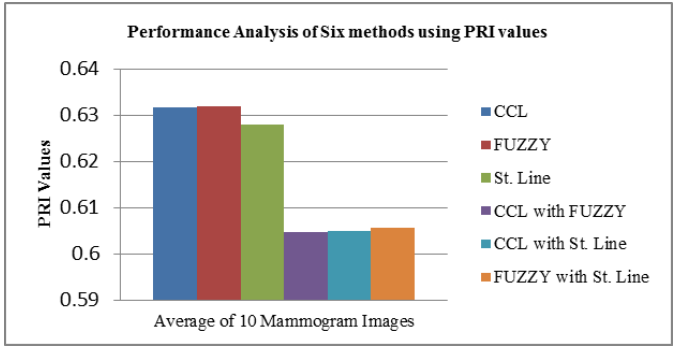

Figure 2. Performance Analysis for PRI values of six methods

It is observed from the Table 2 that proposed methods produces minimized local consistency error values. Figure 3 shows a comparison of all six methods average values of ten mammogram images obtained by using the proposed methods.

TABLE II
LOCAL CONSISTENCY ERROR (LCE) VALUES FOR CCL, FUZZY, STRAIGHT LINE AND PROPOSED METHOD

| MIAS Images | CCL | FUZZY | St. Line |
|---|---|---|---|
| Mdb006 | 0.597624 | 0.59343 | 0.602423 |
| Mdb015 | 0.551666 | 0.548118 | 0.555012 |
| Mdb020 | 0.617434 | 0.616034 | 0.620304 |
| Mdb040 | 0.238447 | 0.236125 | 0.24201 |
| Mdb042 | 0.448288 | 0.445551 | 0.451627 |
| Mdb050 | 0.63335 | 0.631521 | 0.63703 |
| Mdb070 | 0.517977 | 0.511736 | 0.52405 |
| Mdb077 | 0.579295 | 0.575678 | 0.580302 |
| Mdb099 | 0.442121 | 0.439119 | 0.442851 |
| Mdb100 | 0.477111 | 0.474178 | 0.484285 |

| MIAS Images | CCL with FUZZY | CCL with St. Line | FUZZY with St. Line |
|---|---|---|---|
| Mdb006 | 0.594207 | 0.595639 | 0.595499 |
| Mdb015 | 0.55637 | 0.558248 | 0.55791 |
| Mdb020 | 0.660039 | 0.660923 | 0.660321 |
| Mdb040 | 0.259328 | 0.260263 | 0.260085 |
| Mdb042 | 0.459397 | 0.460429 | 0.460218 |
| Mdb050 | 0.677242 | 0.678941 | 0.67808 |
| Mdb070 | 0.578005 | 0.579829 | 0.579347 |
| Mdb077 | 0.576046 | 0.577471 | 0.577408 |
| Mdb099 | 0.448696 | 0.450034 | 0.483309 |
| Mdb100 | 0.482307 | 0.483871 | 0.483309 |

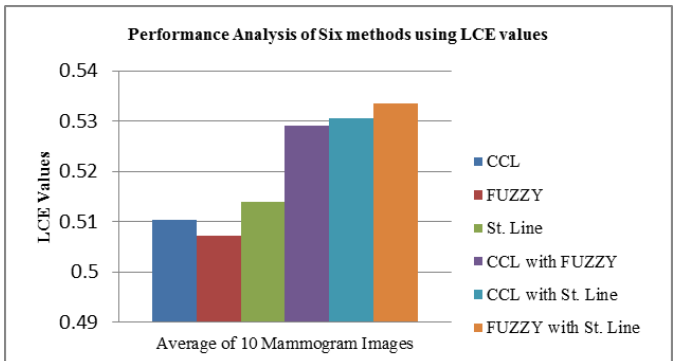

Figure 3. Performance Analysis for LCE values of six methods

It is observed from the Table 3 that proposed methods produces maximized Tanimoto Coefficient values. Figure 4 shows a comparison of all six methods average values of ten mammogram images obtained by using the proposed methods.

TABLE III
TANIMOTO COEFFICIENT (TC) VALUES FOR CCL, FUZZY, STRAIGHT LINE AND PROPOSED METHOD

| MIAS Images | CCL | FUZZY | St. Line |
|---|---|---|---|
| Mdb006 | 0.2 | 0.2 | 0.2 |
| Mdb015 | 0.202522 | 0.204603 | 0.202522 |
| Mdb020 | 0.2 | 0.2 | 0.2 |
| Mdb040 | 0.292834 | 0.293354 | 0.288017 |
| Mdb042 | 0.205158 | 0.206293 | 0.20532 |
| Mdb050 | 0.2 | 0.2 | 0.208754 |
| Mdb070 | 0.209251 | 0.210249 | 0.208754 |
| Mdb077 | 0.2 | 0.2 | 0.2 |
| Mdb099 | 0.201175 | 0.201175 | 0.201175 |
| Mdb100 | 0.210919 | 0.210919 | 0.210584 |

| MIAS Images | CCL with FUZZY | CCL with St. Line | FUZZY with St. Line |
|---|---|---|---|
| Mdb006 | 0.2 | 0.2 | 0.2 |
| Mdb015 | 0.207041 | 0.204603 | 0.206713 |
| Mdb020 | 0.2 | 0.2 | 0.2 |
| Mdb040 | 0.293615 | 0.293354 | 0.293354 |
| Mdb042 | 0.206619 | 0.206293 | 0.206293 |
| Mdb050 | 0.2 | 0.2 | 0.2 |
| Mdb070 | 0.210417 | 0.210249 | 0.210249 |
| Mdb077 | 0.2 | 0.2 | 0.2 |
| Mdb099 | 0.201175 | 0.201175 | 0.201175 |
| Mdb100 | 0.210919 | 0.210919 | 0.210919 |

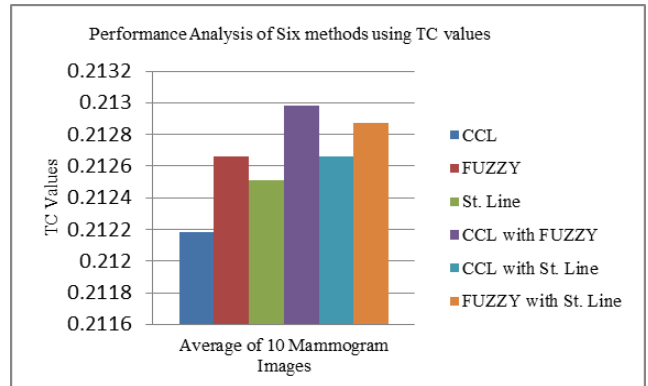

Figure 4. Performance Analysis for TC values of six methods

It is observed from the Table 4 that proposed methods produces minimized mean absolute error values. Figure 5 shows a comparison of all six methods average values of ten mammogram images obtained by using the proposed methods.

TABLE IV
MEAN ABSOLUTE ERROR (MAE) VALUES FOR CCL, FUZZY, STRAIGHT LINE AND PROPOSED METHOD

| MIAS Images | CCL | FUZZY | St. Line |
|---|---|---|---|
| Mdb006 | 14.95815 | 15.13723 | 15.3787 |
| Mdb015 | 7.653979 | 8.021899 | 7.1291 |
| Mdb020 | 11.7442 | 12.04196 | 11.8065 |
| Mdb040 | 4.368523 | 4.577971 | 5.4374 |
| Mdb042 | 18.04244 | 18.19584 | 19.6207 |
| Mdb050 | 3.084916 | 3.418113 | 3.6963 |
| Mdb070 | 2.677366 | 2.922425 | 2.8311 |
| Mdb077 | 17.641 | 17.88212 | 17.9938 |
| Mdb099 | 20.03298 | 20.37581 | 20.7162 |
| Mdb100 | 14.7345 | 15.04974 | 15.9393 |

| MIAS Images | CCL with FUZZY | CCL with St. Line | FUZZY with St. Line |
|---|---|---|---|
| Mdb006 | 16.77865 | 53.13372 | 0.835207 |
| Mdb015 | 41.87288 | 7.482403 | 0.917385 |
| Mdb020 | 40.53586 | 8.710763 | 0.874889 |
| Mdb040 | 43.99408 | 3.567551 | 0.766388 |
| Mdb042 | 31.58742 | 22.74979 | 0.495436 |
| Mdb050 | 62.13871 | 22.39978 | 0.481487 |
| Mdb070 | 27.70708 | 9.981722 | 0.859869 |
| Mdb077 | 18.31366 | 41.58066 | 0.887138 |
| Mdb099 | 55.28955 | 27.3498 | 0.592274 |
| Mdb100 | 43.81568 | 20.04527 | 0.965522 |

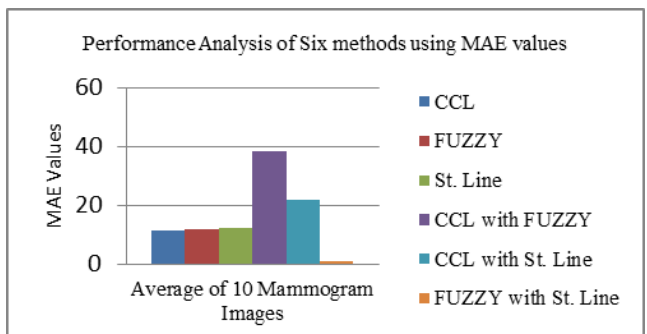

Figure 5. Performance Analysis for MAE values of six methods

It is observed from the Table 5 that proposed methods produces minimized Hausdroff distance values. Figure 6 shows a comparison of all six methods average values of ten mammogram images obtained by using the proposed methods.

TABLE V
HAUSDROFF DISTANCE (HD) VALUES FOR CCL, FUZZY, STRAIGHT LINE AND PROPOSED METHOD

| MIAS Images | CCL | FUZZY | St. Line |
|---|---|---|---|
| Mdb006 | 2807.60 | 2809.00 | 2780.80 |
| Mdb015 | 1997.40 | 1996.00 | 2010.30 |
| Mdb020 | 2877.60 | 2878.00 | 2859.50 |
| Mdb040 | 1423.30 | 1424.00 | 1420.80 |
| Mdb042 | 1973.70 | 1974.00 | 1933.90 |
| Mdb050 | 3043.90 | 3044.00 | 3016.30 |
| Mdb070 | 2442.10 | 2443.00 | 2429.30 |
| Mdb077 | 2526.60 | 2528.00 | 2530.30 |
| Mdb099 | 2240.50 | 2243.00 | 2240.60 |
| Mdb100 | 2220.60 | 2224.00 | 2180.10 |

| MIAS Images | CCL with FUZZY | CCL with St. Line | FUZZY with St. Line |
|---|---|---|---|
| Mdb006 | 3259.30 | 1912.20 | 3144.00 |
| Mdb015 | 2071.80 | 2004.00 | 2133.60 |
| Mdb020 | 2583.90 | 2975.40 | 3176.60 |
| Mdb040 | 2565.20 | 1400.70 | 1487.40 |
| Mdb042 | 6716.90 | 1892.70 | 2342.70 |
| Mdb050 | 1679.00 | 2705.70 | 3147.00 |
| Mdb070 | 1883.20 | 2232.60 | 2503.40 |
| Mdb077 | 1591.90 | 1764.80 | 2895.60 |
| Mdb099 | 1371.60 | 2252.40 | 2727.10 |
| Mdb100 | 1611.80 | 2278.70 | 2577.30 |

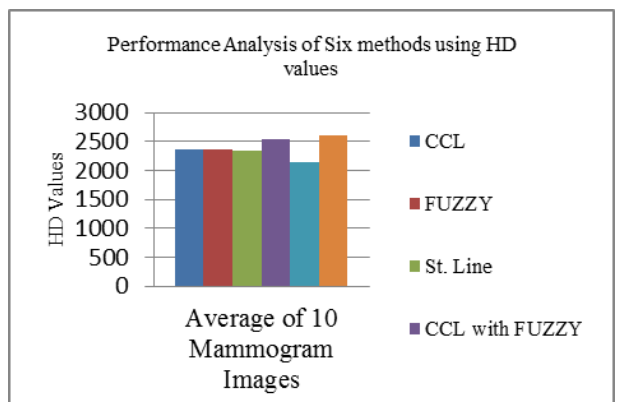

Figure 6. Performance Analysis for HD values of six methods

## III. CONCLUSION

In this paper, an approach to segmentation of the breast region with pectoral muscle removal in mammograms has been proposed based on a hybridization of ccl, fuzzy and straight line methods which are based on the pectoral muscles removal. Initial segmentation results on more than 322 mammograms have been qualitatively evaluated and have shown that our method can robustly obtain an acceptable segmentation in 98.4% and 95.5% for breast-boundary and pectoral muscle separation in mammograms with different density types and preserve the tissue close to the breast skin line effectively.

ACKNOWLEDGMENT

The First author would like to thank the University Grant Commission, India (F. No. 41-1361/2012(SR)) for his research.

**MR. I. LAURENCE AROQUIARAJ** received the M.Sc.,Computer Science from Pondicherry University, Pondicherry, India in 2002, M. Phil., degree from Manonmaniam University, Tamilnadu, India in 2003, Master of Technology from CSE, Kalinga University, India in 2005 and Master of Computer Applications Degree from Periyar University, Salem, Tamilnadu, India in 2009.

He is working as Assistant Professor in Computer Science, Periyar University, Salem, Tamilnadu, India. He is pursuing his Ph.D in medical Image Processing. His research interest includes image processing, biometrics and pattern recognition.

This author became a Member of IEEE in 2012 and also Member of IAENG in 2011.

**Dr. THANGAVEL KUTTIANNAN** received the Master of Science from Department of Mathematics, Bharathidasan University in 1986 and Master of Computer Applications Degree from Madurai Kamaraj University, India in 2001. He obtained his Ph. D. Degree from the Department of Mathematics, Gandhigram Rural University in 1999.

He worked as Reader in the Department of Mathematics, Gandhigram Rural University, upto 2006. Currently he is working as Professor and Head, Department of Computer Science, Periyar University, Salem, Tamilnadu, India. His areas of interest include medical image processing, artificial intelligence, neural network, fuzzy logic, data mining, pattern recognition and mobile computing.

This author became a Member of IEEE in 2012. He was received Young Scientist award, Tamilnadu Government, India in 2010.